\documentclass[12pt,onecolumn]{IEEEtran}
\usepackage{booktabs}
\usepackage{multirow}
\usepackage{cite}
\usepackage{enumitem}
\usepackage{graphicx}
\usepackage{textcomp}
\usepackage{xcolor}
\usepackage{float}
\usepackage{microtype}
\usepackage{makecell}
\usepackage{subcaption}
\usepackage{wrapfig}
\usepackage{bbding}
\usepackage{amsmath,amssymb,amsfonts}
\usepackage{soul}
\usepackage{algorithmic}
\usepackage{cite}
\usepackage{amsmath,amssymb,amsfonts}
\usepackage{algorithmic}
\usepackage{graphicx}
\usepackage{textcomp}
\usepackage{xcolor}
\def\BibTeX{{\rm B\kern-.05em{\sc i\kern-.025em b}\kern-.08em
    T\kern-.1667em\lower.7ex\hbox{E}\kern-.125emX}}

\begin{document}
\title{HGC-Herd: Efficient Heterogeneous Graph Condensation via Representative Node Herding\\
{\footnotesize \textsuperscript{}}
}
\author{\IEEEauthorblockN{1\textsuperscript{st} Fuyan Ou }
\IEEEauthorblockA{\textit{College of Computer and Information Science} \\
\textit{Southwest University}\\
Chongqing, China \\
omaksimovbe@gmail.com}
\and
\IEEEauthorblockN{2\textsuperscript{nd} Siqi Ai}
\IEEEauthorblockA{\textit{College of Computer and Information Science} \\
\textit{Southwest University}\\
Chongqing, China  \\
599900503@qq.com}
\and
\IEEEauthorblockN{1\textsuperscript{st} Yulin Hu\textsuperscript{*}}
\IEEEauthorblockA{\textit{College of Computer and Information Science} \\
\textit{Southwest University}\\
Chongqing, China \\
\textsuperscript{*}hylinserein@gmail.com}
}
\maketitle

\begin{abstract}
Heterogeneous graph neural networks (HGNNs) have demonstrated strong capability in modeling complex semantics across multi-type nodes and relations. However, their scalability to large-scale graphs remains challenging due to structural redundancy and high-dimensional node features. Existing graph condensation approaches, such as GCond, are primarily developed for homogeneous graphs and rely on gradient matching, resulting in considerable computational, memory, and optimization overhead.  
We propose \textbf{HGC-Herd}, a training-free condensation framework that generates compact yet informative heterogeneous graphs while maintaining both semantic and structural fidelity. HGC-Herd integrates lightweight feature propagation to encode multi-hop relational context and employs a class-wise herding mechanism to identify representative nodes per class, producing balanced and discriminative subsets for downstream learning tasks.  
Extensive experiments on ACM, DBLP, and Freebase validate that HGC-Herd attains comparable or superior accuracy to full-graph training while markedly reducing both runtime and memory consumption. These results underscore its practical value for efficient and scalable heterogeneous graph representation learning.
\end{abstract}
\begin{IEEEkeywords}
graph neural network, heterogeneous graph, node classification, heterogeneous graph condensation
\end{IEEEkeywords}

\section{Introduction}
Heterogeneous graphs in recommendation, academic, and social applications provide rich semantics but also make scalable representation learning challenging. Their multiple node and edge types and complex semantic–structural dependencies render HGNNs computationally heavy, and models like HAN~\cite{HAN} and Simple-HGN~\cite{SimpleHGN} rely on repeated aggregation and high-dimensional propagation. Enlarged neighborhoods further increase receptive fields and memory usage, limiting deployment under constrained resources. As metapaths and relation types grow, mini-batch training often requires sampling and caching, adding extra overhead.

Sampling-based methods (e.g., FastGCN, GraphSAINT~\cite{Fast-GCN,GraphSAINT}) reduce computation but introduce information loss and structural bias. Extending them to heterogeneous settings is difficult because metapath semantics and cross-type neighborhoods must be preserved; naive sampling may under-cover key relations and disturb class distributions. These limitations motivate \textit{data-level condensation}, which builds compact graphs that maintain core semantics for efficient and reliable HGNN training.

We propose \textbf{HGC-Herd}, a \textit{training-free} heterogeneous graph condensation method using prototype-guided representative node selection. Instead of gradient optimization, HGC-Herd performs lightweight feature propagation across meta-relations~\cite{RGCN} and applies class-wise herding to choose nodes near class prototypes. This design preserves semantic diversity and structural consistency while significantly reducing graph size. The decoupled procedure avoids bi-level sensitivity and supports controllable class-wise budgets and stable selection across random seeds.

HGC-Herd generates compact subgraphs that retain essential semantics and cross-type connectivity. Training HGNNs such as Simple-HGN~\cite{SimpleHGN} on the condensed graphs achieves accuracy close to full-graph training with much lower computation and memory. Experiments on ACM~\cite{HAN}, DBLP~\cite{HGB}, and Freebase~\cite{Freebase} demonstrate efficiency, scalability, and generalization, with ablation studies confirming the effectiveness of propagation and herding. Notable reductions in wall-clock time and GPU usage further highlight the practicality of HGC-Herd. Our key contributions are:
\begin{itemize}[leftmargin=1.2em]
    \item A \textbf{training-free} condensation framework with prototype-guided node selection, providing reproducibility and seamless HGNN integration.
    \item A \textbf{class-wise herding} strategy that preserves semantic diversity, supports flexible budgets, and maintains cross-type structure.
    \item Competitive accuracy with only $1.2\%$ of data, outperforming sampling and gradient-based baselines with lower runtime and memory overhead.
\end{itemize}

\section{Related Work}
\noindent\textbf{Heterogeneous Graph Neural Networks.}
HGNNs extend GNNs to handle multi-type nodes and relations, enabling expressive modeling in recommendation, citation, and knowledge graphs. Representative models such as R-GCN~\cite{RGCN}, HAN~\cite{HAN}, and Simple-HGN~\cite{SimpleHGN} capture relation-specific semantics via attention or meta-path aggregation.~\cite{10751750,10804824} However, training remains costly due to inter-type interactions and repeated propagation. Efficiency-oriented techniques must preserve type semantics and metapath structures, while choices of metapaths, depth, and sampling introduce sensitivity, motivating data-level methods with stronger semantic guarantees.~\cite{10.1145/3711896.3737126}

\noindent\textbf{Graph Reduction.}
Graph reduction lowers GNN cost by generating smaller yet informative graphs. Common strategies include coreset selection~\cite{herding}, sparsification~\cite{sparsification1}, and coarsening~\cite{coarseninghuang}. Although effective, extreme compression ($>99\%$) often harms performance and distorts semantics, motivating condensation techniques that better preserve distributions. Many reduction methods are topology-centered and fail to ensure label balance or metapath coverage, both crucial for heterogeneous graphs.~\cite{10400813,10159989,9072622}

\noindent\textbf{Graph Condensation.}
Graph condensation synthesizes compact graphs retaining key structure and features. GCond~\cite{gcond} matches gradients between real and synthetic graphs, while SGDD~\cite{sgdd}, GCSR~\cite{gcsr}, and SFGC~\cite{sfgc} introduce structural priors or trajectory alignment. Distribution-based methods (e.g., GCDM~\cite{gcdm}) improve fidelity but rely on bi-level optimization, which is hard to scale and demands extensive tuning.

\noindent\textbf{Comparison and Motivation.}
Most condensation methods focus on homogeneous graphs and depend heavily on optimization, limiting applicability to heterogeneous settings. \textbf{HGC-Herd} introduces a \textit{training-free}, prototype-guided paradigm that combines lightweight feature propagation with class-wise herding to preserve structural and semantic diversity efficiently. It achieves strong results on ACM~\cite{HAN}, DBLP~\cite{HGB}, and Freebase~\cite{Freebase}. The framework emphasizes simplicity and determinism: once the budget is defined, selection is stable and reproducible, offering a practical data-side reduction scalable to multi-type graphs without modifying HGNN architectures.~\cite{9839318}

\begin{table*}[t]
\renewcommand\arraystretch{1.2}
\caption{\textbf{Node classification accuracy (\%) on ACM, DBLP, and Freebase under different condensation ratios.}}
\centering
\resizebox{0.85\textwidth}{!}{
\begin{tabular}{*{8}{c}}
\toprule
& &\multicolumn{4}{c}{Baselines}& Proposed\\
\cmidrule(r){3-6} \cmidrule(r){7-7}
Dataset & Ratio (r) & Random-HG & K-Center-HG & Coarsening-HG & GCond & \pmb{HGC-Herd} & Whole Dataset\\
\hline
\multirow{4}{*}{ACM}
&1.2\%&$53.37\pm0.24$&$62.66\pm0.26$&$64.17\pm0.31$&$41.17\pm5.14$&$\pmb{91.88\pm0.51}$&\multirow{4}{*}{$93.11\pm0.62$}\\
&2.4\%&$61.18\pm0.36$&$65.45\pm1.21$&$65.56\pm0.24$&$46.07\pm6.86$&$\pmb{93.01\pm0.32}$&\\
&4.8\%&$60.01\pm2.36$&$69.68\pm0.76$&$68.91\pm0.73$&$56.12\pm3.42$&$\pmb{92.07\pm0.47}$&\\
&9.6\%&$66.25\pm1.18$&$75.68\pm0.76$&$70.91\pm0.73$&$65.56\pm6.15$&$\pmb{93.01\pm0.21}$&\\
\hline
\multirow{4}{*}{DBLP}
&1.2\%&$38.73\pm0.98$&$61.39\pm0.26$&$53.27\pm0.22$&$53.26\pm7.26$&$\pmb{89.86\pm0.35}$&\multirow{4}{*}{$95.19\pm0.31$}\\
&2.4\%&$48.84\pm2.36$&$63.80\pm0.78$&$59.63\pm0.65$&$51.13\pm8.25$&$\pmb{93.59\pm0.25}$&\\
&4.8\%&$45.49\pm0.52$&$70.68\pm0.37$&$66.21\pm0.25$&$57.49\pm7.54$&$\pmb{94.01\pm0.34}$&\\
&9.6\%&$56.01\pm2.36$&$79.68\pm0.76$&$76.91\pm0.73$&$64.25\pm2.77$&$\pmb{94.79\pm0.26}$&\\
\hline
\multirow{4}{*}{Freebase}
&1.2\%&$45.32\pm0.47$&$48.18\pm0.26$&$46.28\pm0.56$&$51.24\pm0.45$&$\pmb{57.18\pm0.17}$&\multirow{4}{*}{$63.67\pm0.34$}\\
&2.4\%&$47.52\pm0.14$&$48.85\pm0.71$&$49.10\pm0.56$&$52.34\pm0.56$&$\pmb{60.17\pm0.31}$&\\
&4.8\%&$48.15\pm1.21$&$51.33\pm0.86$&$50.25\pm0.56$&$55.81\pm2.31$&$\pmb{61.40\pm0.23}$&\\
&9.6\%&$50.01\pm2.36$&$52.68\pm0.76$&$52.91\pm0.73$&$57.03\pm0.23$&$\pmb{61.42\pm0.18}$&\\
\hline
\end{tabular}}
\label{tab:main_results}
\end{table*}

\section{HGC-Herd Framework}
\begin{figure}[t!]
    \centering
    % 占满正文宽度
    \includegraphics[width=\linewidth]{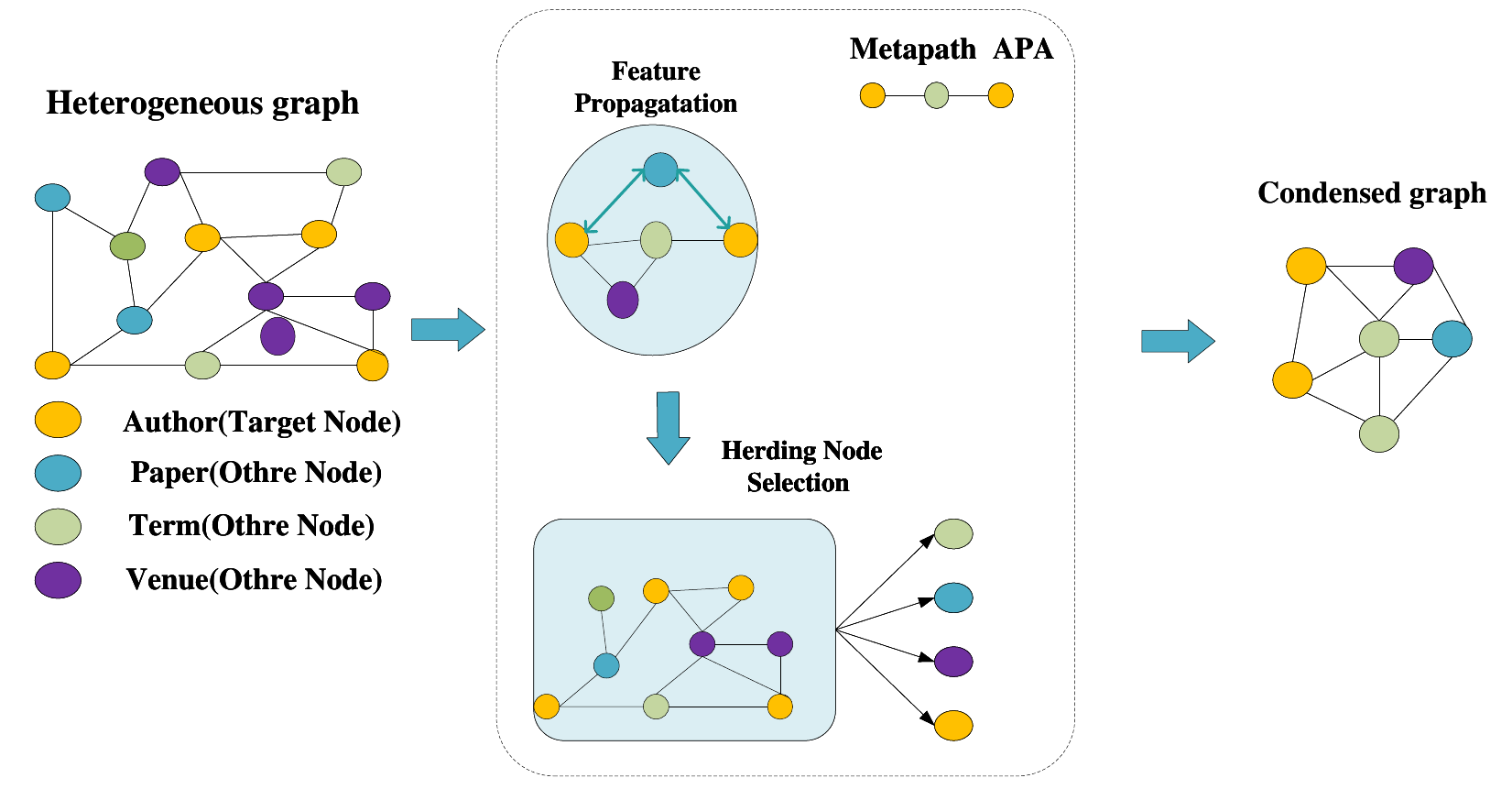}
    % 或者稍微留一点空白：0.9\textwidth
    % \includegraphics[width=0.9\textwidth]{HGC_Herd/picture/framework.pdf}
    \caption{The workflow of HGC-Herd.}
    \label{fig:framework}
\end{figure}

\textbf{HGC-Herd} condenses heterogeneous graphs into compact, semantics-preserving subgraphs for efficient HGNN training. As shown in Figure~\ref{fig:framework}, it comprises three key modules: (1) \textit{Feature Propagation} for metapath-based aggregation; (2) \textit{Class-wise Prototype Construction} for forming class-level centroids; and (3) \textit{Strategic Herding Selection} for selecting nodes that best represent class distributions. Collectively, these components form a fully \textit{training-free} pipeline that maintains both semantic and structural integrity while substantially reducing graph size, enabling HGNNs to achieve comparable accuracy with far lower computational cost. The modular design also allows the propagation operator and selection budget to adapt to new tasks without retraining the condensation pipeline.

\subsection{Feature Propagation}
Feature propagation serves as a preprocessing stage that integrates structural and semantic information. Unlike HGNNs that aggregate neighbors at every epoch, SeHGNN performs propagation once prior to training, markedly reducing computation. Given a heterogeneous graph, propagation follows metapaths $\mathcal{P}=c c_1 c_2\dots c_l$, where $c$ is the target node type. The aggregated representation along $\mathcal{P}$ is defined as
\begin{equation}
X^{\mathcal{P}} = \hat{A}_{c,c_1}\hat{A}_{c_1,c_2}\dots\hat{A}_{c_{l-1},c_l}X^{c_l},
\label{eq:metapath}
\end{equation}
where $\hat{A}_{c_i,c_{i+1}}$ denotes the row-normalized adjacency between types $c_i$ and $c_{i+1}$, and $X^{c_l}$ represents the feature matrix of nodes of type $c_l$. This operation aggregates multi-hop semantics along heterogeneous paths, enabling precomputation and reuse of intermediates (e.g., $X^{PAP}$) while avoiding redundant attention. The fused features are later employed for downstream HGNN training, consistent with HGC-Herd’s training-free design. In practice, we cache $X^{\mathcal{P}}$ for frequently used metapaths to amortize preprocessing and ensure reproducibility.

\subsection{Class-wise Prototype Construction}
After propagation, HGC-Herd computes class-level prototypes for target-type nodes (e.g., papers). For each class $c$,
\begin{equation}
\mu_c = \frac{1}{|V_c|}\sum_{v \in V_c}\mathbf{h}_v',
\label{eq:proto}
\end{equation}
where $\mathbf{h}_v'$ denotes the propagated feature from Eq.~\ref{eq:metapath}. The prototype $\mu_c$ captures the central semantic representation of class $c$ and guides representative node selection. To enhance stability, prototypes are computed using all labeled nodes of the target type. Under class imbalance, the selection budget for each class is proportional to $|V_c|$, ensuring balanced representation and preventing dominant classes from overwhelming minority ones.

\subsection{Strategic Herding Selection}
Given the computed class-wise prototypes $\{\mu_c\}$, HGC-Herd adopts a greedy herding strategy to select representative nodes. For each class $c$, let $V_c$ be the candidate node set and $S_c$ the currently selected subset. At each step we choose
\begin{equation}
v_i^* = \arg\min_{v_i \in V_c \setminus S_c}
\left\|
\mu_c - \frac{1}{|S_c \cup \{v_i\}|}
\sum_{u_j \in S_c \cup \{v_i\}} \mathbf{h}_j'
\right\|_2,
\label{eq:herding}
\end{equation}
where $\mathbf{h}_j'$ denotes the propagated feature of node $u_j$ and $\|\cdot\|_2$ is the Euclidean norm. Starting from $S_c=\emptyset$, this process iteratively adds nodes until the budget $b_c$ (or ratio $r_c$) is reached, yielding per-class ``herds'' that approximate $\mu_c$ in feature space. The selection is deterministic under a fixed ordering, and its complexity grows linearly with the number of candidates per class, making the condensation scalable to large heterogeneous graphs.

\section{Experiments}
We evaluate \textbf{HGC-Herd} on three representative heterogeneous graph benchmarks to assess its efficiency and effectiveness. We evaluate node classification on the designated target type for each dataset and analyze both predictive quality and computational savings.

\subsection{Datasets}
Experiments are conducted on ACM, DBLP, and Freebase. ACM is a bibliographic network with papers, authors, and subjects; DBLP captures co-authorship and citation relations among authors, papers, and venues; Freebase is a large-scale knowledge graph containing multiple entity types (e.g., books, fields, persons). For each dataset, node classification is performed on the target node type (\textit{paper}, \textit{author}, \textit{book}) following standard splits. We preserve the original features and edge types without external pretraining. The dataset statistics are summarized in Table~\ref{tab:Datasets}.

\begin{table}[t!]
\renewcommand\arraystretch{1}
\caption{\textbf{Dataset statistics.}}
\setlength{\tabcolsep}{1mm}
\centering
\resizebox{0.45\textwidth}{!}{
\begin{tabular}{*{7}{c}}
\hline
\makecell[c]{Dataset} & \#Nodes & \makecell[c]{\#Node\\types} & \#Edges & \makecell[c]{\#Edge\\types} & Target & \#Classes \\
\hline
DBLP & 26{,}128 & 4 & 239{,}566 & 6 & author & 4 \\
ACM & 10{,}942 & 4 & 547{,}872 & 8 & paper & 3 \\
Freebase & 180{,}098 & 8 & 1{,}057{,}688 & 36 & book & 7 \\
\hline
\end{tabular}}
\label{tab:Datasets}
\vspace{0.6em}
\end{table}

\subsection{Baselines and Implementation Details}
We compare against: (1) \textbf{Random-HG} (random node sampling); (2) \textbf{K-Center-HG} (greedy K-center coverage); (3) \textbf{Coarsening-HG} (METIS-style structural coarsening); (4) \textbf{GCond} (gradient-based condensation for homogeneous graphs); and (5) \textbf{Whole Dataset} (upper bound).  
All methods adopt \textbf{Simple-HGN} with 64 hidden units, two layers, and Adam ($lr=5\times10^{-4}$). We evaluate condensation ratios $r\!\in\!\{1.2\%,2.4\%,4.8\%,9.6\%\}$, repeat each run five times on an NVIDIA RTX 3050 GPU, and report mean accuracy (\%). Early stopping is applied based on validation accuracy, and random seeds are fixed for reproducibility. Condensed graphs are constructed once per ratio and reused across all training runs.

\begin{figure}[t!]
\centering
\includegraphics[width=3.3in]{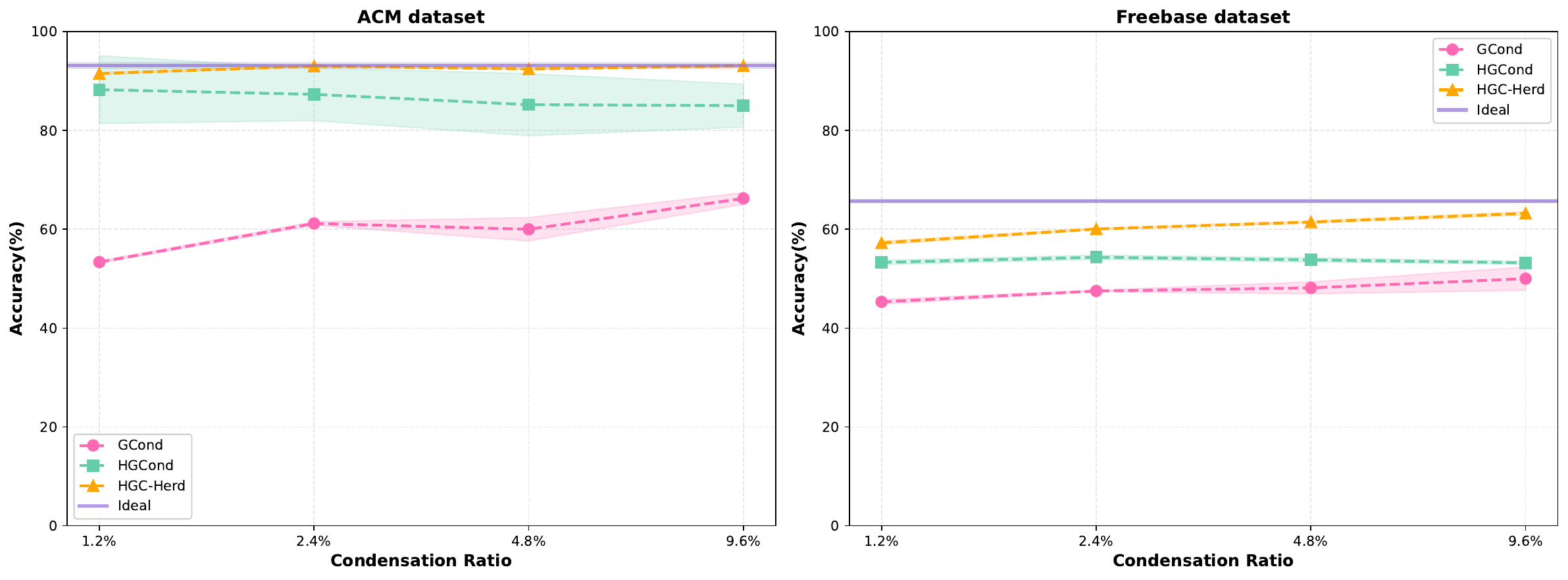}
\caption{Performance under different condensation ratios.}
\label{fig:efficiency}
\vspace{-0.5cm}
\end{figure}

\begin{figure}[t!]
\centering
\includegraphics[width=3.3in]{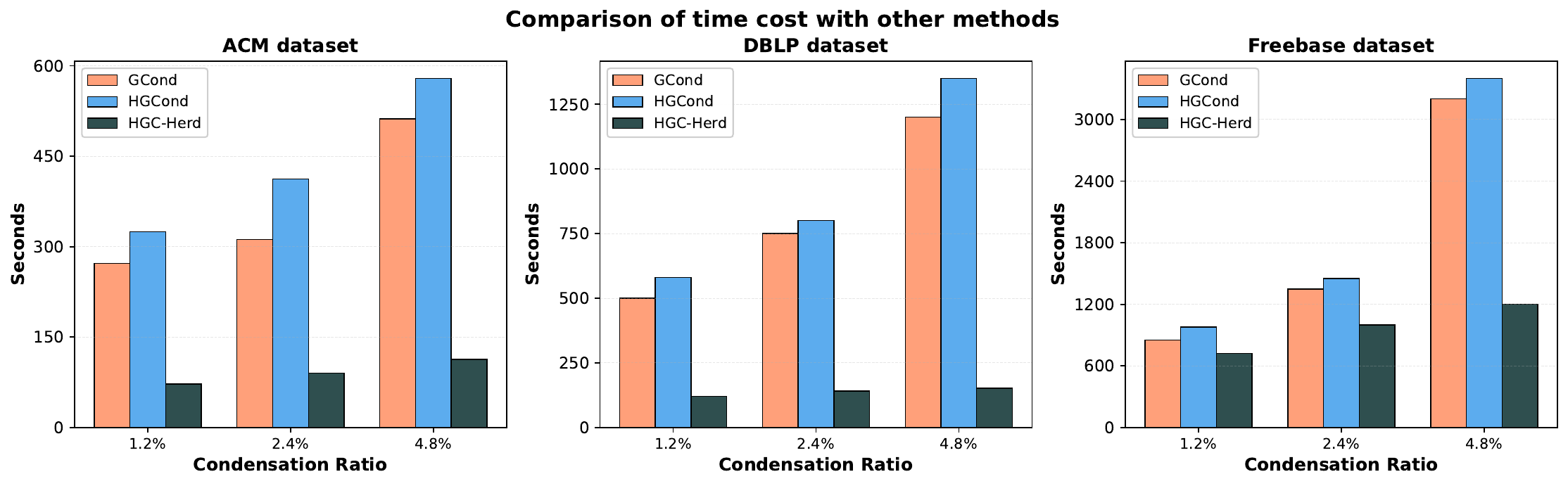}
\caption{Runtime comparison with other methods.}
\label{fig:time_cost_analysis}
\vspace{-0.5cm}
\end{figure}

\subsection{Main Results}
Table~\ref{tab:main_results} reports node classification accuracy across ACM, DBLP, and Freebase under different condensation ratios. \textbf{HGC-Herd} achieves high and stable accuracy, approaching full-graph training even with aggressive compression. On \textbf{ACM}, HGC-Herd attains $91.88\%$ at $1.2\%$ of data and $93.01\%$ at $9.6\%$, nearly matching the $93.11\%$ full-graph result, indicating that class-aware selection effectively preserves semantic structure. On \textbf{DBLP}, it reaches $89.86\%$ at $1.2\%$ and $94.79\%$ at $9.6\%$ (full graph $95.19\%$), demonstrating strong retention of author-topic signals. On the larger and more complex \textbf{Freebase}, it achieves $57.18\%$ at $1.2\%$ and $61.42\%$ at $9.6\%$, close to the $63.67\%$ full-graph accuracy. The relative improvement over random and heuristic baselines is especially pronounced at low budgets, where preserving class prototypes is most critical.

\subsection{Efficiency Analysis}
Figures~\ref{fig:efficiency} and \ref{fig:time_cost_analysis} show that \textbf{HGC-Herd} achieves the lowest condensation cost and stable accuracy across datasets. As seen in Fig.~\ref{fig:efficiency}, its performance remains close to the full-graph baseline and consistently exceeds that of GCond and HGCond, particularly under small condensation ratios such as $1.2\%$. 
In Fig.~\ref{fig:time_cost_analysis}, HGC-Herd achieves about $4$–$6\times$ faster condensation compared with gradient-based methods while maintaining comparable accuracy. This efficiency benefits from its training-free design, which avoids gradient computation and allows the condensed graphs to be reused across runs. Overall, these results confirm that HGC-Herd provides a practical and scalable solution for efficient HGNN training.

\begin{table}[t!]
\renewcommand\arraystretch{1.1}
\caption{\textbf{Ablation study (accuracy \%) at 1.2\% ratio.}}
\centering
\begin{tabular}{lcc}
\toprule
Variant & ACM & Freebase \\
\midrule
HGC-Herd (full) & $\pmb{91.88\pm0.51}$ & $\pmb{57.18\pm0.17}$ \\
w/o Feature Propagation & $89.45\pm0.67$ & $55.12\pm0.89$ \\
w/o Herding & $81.23\pm0.45$ & $48.67\pm0.72$ \\
\bottomrule
\end{tabular}
\label{tab:ablation}
\end{table}

\subsection{Ablation Study}
We evaluate the contribution of each component on ACM and Freebase at $1.2\%$.  
\textbf{w/o Feature Propagation} removes Eq.~\ref{eq:metapath} and directly uses raw features;  
\textbf{w/o Herding} selects top-$k$ nodes closest to prototypes without Eq.~\ref{eq:herding}.  
Table~\ref{tab:ablation} shows that both modules are essential. Removing propagation weakens semantic context, while removing herding reduces coverage and class balance, confirming that both stages jointly ensure faithful and efficient condensation.

\section{Conclusion}
We presented HGC-Herd, a training-free framework for heterogeneous graph condensation that combines feature propagation with strategic herding to produce compact and representative subgraphs. The method achieves a strong balance between efficiency and performance, outperforming competitive baselines on key benchmarks. In future work, we plan to extend HGC-Herd to dynamic graphs, explore adaptive per-class budgets, and investigate integration with federated and privacy-aware settings, where training-free condensation can provide additional system-level advantages.

% 确保 23–72 号文献全部出现在参考文献列表中
\nocite{He2024,Wu2024a,Wu2024b,Chen2024a,Wu2023a,Tang2025,Lin2025,Yang2025,%
Luo2024a,Liao2024,Wu2024c,Chen2024b,Chen2024c,Yang2024,Zhong2024,Yuan2024a,%
Chen2024d,Zeng2024,Yuan2024b,Li2024,Wu2024d,Qin2024a,Li2024b,Bi2024a,%
Qin2024b,Li2024c,Liu2024,Wu2024e,Chen2024e,Jin2024,Xie2024,Qin2024c,Yan2024,%
Wei2024,Jiang2024,Chen2024f,Wang2024,Li2024d,Luo2023a,Luo2023b,Luo2023c,%
Luo2023d,Wu2023b,Chen2023,Bi2023a,Hu2023,Bi2023b,Yang2023,Liu2023}

\bibliographystyle{IEEEtran}
\bibliography{reference}
\end{document}